\pdfoutput=1

\documentclass[11pt]{article}

\usepackage[preprint]{acl}

\usepackage{times}
\usepackage{latexsym}

\usepackage[T1]{fontenc}

\usepackage[utf8]{inputenc}

\usepackage{microtype}

\usepackage{inconsolata}

\usepackage{graphicx}

\usepackage{amsmath}
\usepackage[customcolors]{hf-tikz}

\usepackage{multirow}
\usepackage{listings}
\usepackage{color}
\usepackage{subcaption}
\usepackage{makecell}
\usepackage{booktabs}
\usepackage{cleveref}
\usepackage{colortbl}
\usepackage{soul}
\usepackage{xcolor}
\usepackage{empheq}
\usepackage[many]{tcolorbox}
\usepackage{stfloats}
\usepackage{bbding}
\usepackage{tabularx}

\definecolor{dkgreen}{rgb}{0,0.6,0}
\definecolor{gray}{rgb}{0.5,0.5,0.5}
\definecolor{mauve}{rgb}{0.58,0,0.82}

\definecolor{dgreen}{rgb}{0.412,0.741,0.271}
\definecolor{dblue}{rgb}{0.220,0.325,0.639}
\definecolor{dred}{rgb}{0.933,0.122,0.137}

\definecolor{g1}{HTML}{b3e2cd}
\definecolor{r1}{HTML}{fdcdac}
\definecolor{w1}{HTML}{cbd5e8}
\definecolor{b1}{HTML}{fff7bc}

\definecolor{lr}{HTML}{bebada}
\definecolor{fr}{HTML}{fccde5}

\definecolor{Lavender}{HTML}{BF94E4}

\newcommand{\ie}{\textit{i}.\textit{e}.\,}
\newcommand{\eg}{\textit{e}.\textit{g}.\,}

\definecolor{l1}{RGB}{189,215,238}
\definecolor{l2}{RGB}{222,235,247}
\definecolor{l3}{RGB}{255,230,153}
\definecolor{l4}{RGB}{248,203,173}
\definecolor{l5}{RGB}{244,177,131}

\newtcbox{\columntcbox}{highlight math style={
        colback=gray!30,
        arc=2pt,
        outer arc=2pt,
        boxrule=0pt,
        top=2pt,
        bottom=2pt,
        left=2pt,
        right=2pt,
    }
}

\colorlet{LightLavender}{Lavender!35!}
\newtcbox{\inlinetcbox}[1][]{on line, 
        boxsep=2pt, left=0pt,right=0pt,top=0pt,bottom=0pt,
        colframe=white,colback=LightLavender,  
        highlight math style={enhanced}, #1
}

\newcommand{\inlinetcboxmathsuperscriptonly}[3][]{\inlinetcbox[#1]{{#2}$^{#3}$}}

\newcolumntype{C}[1]{>{\centering}m{#1}}

\makeatletter
\newcommand{\xMapsto}[2][]{\ext@arrow 0599{\Mapstofill@}{#1}{#2}}
\def\Mapstofill@{\arrowfill@{\Mapstochar\Relbar}\Relbar\Rightarrow}
\makeatother

\newcommand{\ourmethod}{\texttt{R$^3$Mem}}

\title{\ourmethod{}: Bridging Memory Retention and Retrieval \\via Reversible Compression}

\author{
    Xiaoqiang Wang\textsuperscript{\rm 1,2}, Suyuchen Wang\textsuperscript{\rm 1,2}, Yun Zhu\textsuperscript{\rm 3}\thanks{Work done while at Google.} \and Bang Liu\textsuperscript{\rm 1,2,4}\thanks{Corresponding author.} \\
    \textsuperscript{\rm 1}DIRO \& Institut Courtois, Universit{\'e} de Montr{\'e}al \\
    \textsuperscript{\rm 2}Mila - Quebec AI Institute; \textsuperscript{\rm 3}Google; \textsuperscript{\rm 4}Canada CIFAR AI Chair \\
    \{\texttt{xiaoqiang.wang, suyuchen.wang, bang.liu}\}\texttt{@umontreal.ca}, \texttt{yunzhu@google.com}
}

\begin{document}
\maketitle
\begin{abstract}

Memory plays a key role in enhancing LLMs' performance when deployed to real-world applications.
Existing solutions face trade-offs: explicit memory designs based on external storage require complex management and incur storage overhead, while implicit memory designs that store information via parameters struggle with reliable retrieval.
In this paper, we propose \textbf{\ourmethod{}}, a memory network that optimizes both information \textbf{R}etention and \textbf{R}etrieval through \textbf{R}eversible context compression. 
Specifically, \ourmethod{} employs virtual memory tokens to compress and encode infinitely long histories, further enhanced by a hierarchical compression strategy that refines information from document- to entity-level for improved assimilation across granularities.
For retrieval, \ourmethod{} employs a reversible architecture, reconstructing raw data by invoking the model backward with compressed information. Implemented via parameter-efficient fine-tuning, it can integrate seamlessly with any Transformer-based model.
Experiments demonstrate that our memory design achieves state-of-the-art performance in long-context language modeling and retrieval-augmented generation tasks. It also significantly outperforms conventional memory modules in long-horizon interaction tasks like conversational agents, showcasing its potential for next-generation retrieval systems.

\end{abstract}

\section{Introduction}
\label{sec:introduction}

Large language models (LLMs)~\citep{ouyang2022training,team2023gemini,dubey2024llama} have demonstrated remarkable capabilities in natural language understanding and generation~\citep{liang2022holistic,srivastava2023beyond,wang-etal-2024-fac2e}, achieving human-comparable performance on complex reasoning tasks~\citep{guo2023close,suzgun2024meta}. Deploying LLMs as controllers to interact with dynamic environments and solve real-world tasks, \ie, as autonomous agents, has shown promising success across diverse applications, including conversational assistants~\citep{chatgpt, achiam2023gpt}, workflow automation~\citep{hong2023metagpt,wu2024copilot,wang2024oscar,qin2025ui}, and embodied navigation~\citep{wang2023voyager,zheng2024towards,sun-etal-2024-enhancing-agent}. 

However, LLMs have inherent limitations: their stateless nature~\citep{sumers2023cognitive} makes them struggle with leveraging past experiences for multi-turn interactions and cross-task generalization. Furthermore, their reliance on fixed context windows and static parameterized knowledge constrains their ability to handle complex tasks requiring dynamic, up-to-date information~\citep{tao2024survey}.


\begin{figure}[!t]
\centering
    \includegraphics[width=1.0\columnwidth]{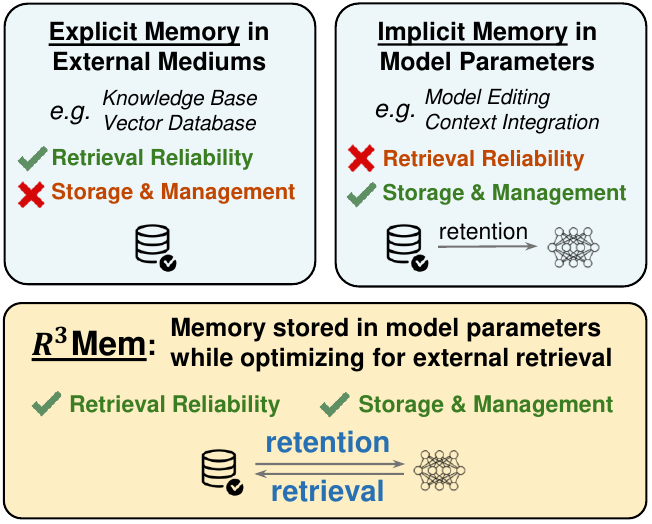}
    \caption{Comparison between explicit memory, implicit memory, and our proposed \ourmethod{} memory design.}
    \label{fig:motivation-example}
    \vspace{-5mm}
\end{figure}

To address these challenges, existing approaches introduce external storage (\ie, explicit memory), such as knowledge repositories~\citep{kagaya2024rap,zhu2024knowagent} and vector databases~\citep{liu2024retrievalattention,jing2024large}, to enhance long-term retention and enable cross-task generalization~\citep{maharana2024evaluating,wang2023voyager,wang2023jarvis} and cross-model sharing~\citep{gao2024memory}. 
In parallel, implicit memory encodes contextual information directly into model parameters, enabling continuous knowledge updates while providing a more compact representation of information, reducing redundancy compared to external storage.
Model-editing methods modify neurons to update~\citep{huang2023transformer,gangadhar-stratos-2024-model} or forget knowledge~\citep{wang2024large}, while context integration~\citep{choi2022prompt,wang2024self} adjusts internal parameters via model distillation. Memory-augmented Transformers (\eg, RMT~\citep{bulatov2022recurrent}, Associate Memory~\citep{he2024camelot,wang2024memoryllm,tack2024online}, and Titans~\citep{behrouz2024titans}) enhance retention by integrating dedicated memory components.

However, as illustrated in Figure~\ref{fig:motivation-example}, both explicit and implicit memory involve trade-offs between storage overhead and recall effectiveness. Explicit memory grows indefinitely, requiring complex memory management techniques such as merging~\citep{yin-etal-2024-explicit,hu2024hiagent} and forgetting~\citep{zhong2024memorybank}. In contrast, implicit memory suffers from unreliable retrieval due to the black-box nature of LLMs, leading to confabulation and hallucination issues~\citep{li2024banishing}. As analyzed by \citet{padmanabhan2024propagating}, injected atomic facts can propagate and influence broader inferences, further complicating retrieval accuracy. More recently, adaptive retrieval~\citep{mallen-etal-2023-trust,farahani2024deciphering} and MemoRAG~\citep{qian2024memorag} combine explicit and implicit memory in a hybrid retrieval paradigm but remain dependent on large-scale external storage.

In this paper, we propose \textbf{\ourmethod{}}, a novel memory-augmented model that optimizes both \underline{\textbf{mem}}ory \underline{\textbf{r}}etention and \underline{\textbf{r}}etrieval while minimizing external storage dependency. \ourmethod{} leverages a \underline{\textbf{r}}eversible architecture that integrates context compression and expansion, enabling assimilation and reconstruction of input data.

Specifically, we design a context compression task that learns to generate compressed representations (`query') from raw input (`context'). \ourmethod{} utilizes virtual memory tokens to encode and retain text that is indefinitely long. To improve compression quality, we introduce a hierarchical compression strategy, progressively refining information at the document, paragraph, and entity levels. 

For retrieval, \ourmethod{} adopts a reversible architecture, reconstructing raw input by inverting the model invocation on compressed representations. This is achieved through adapter tuning, allowing seamless integration with pre-trained Transformer model while maintaining parameter efficiency.

To optimize both memory retention and retrieval, we employ bidirectional training with cycle consistency. The forward process encodes context into compressed memory representations, while the backward process reconstructs the raw content from memory tokens, enforcing consistency between the original and reconstructed information.

We evaluate \ourmethod{} on memory-intensive tasks, achieving state-of-the-art performance in long-context language modeling and retrieval-augmented generation. We also integrate \ourmethod{} into a real-world conversational agent that requires long-horizon interactions and the ability to recall distant historical context. \ourmethod{} consistently outperforms existing memory modules, demonstrating superior scalability, retrieval accuracy, and potential for next-generation retrieval systems.

\section{Methodology}
\label{sec:methodology}

\begin{figure*}[!t]
    \includegraphics[width=\textwidth]{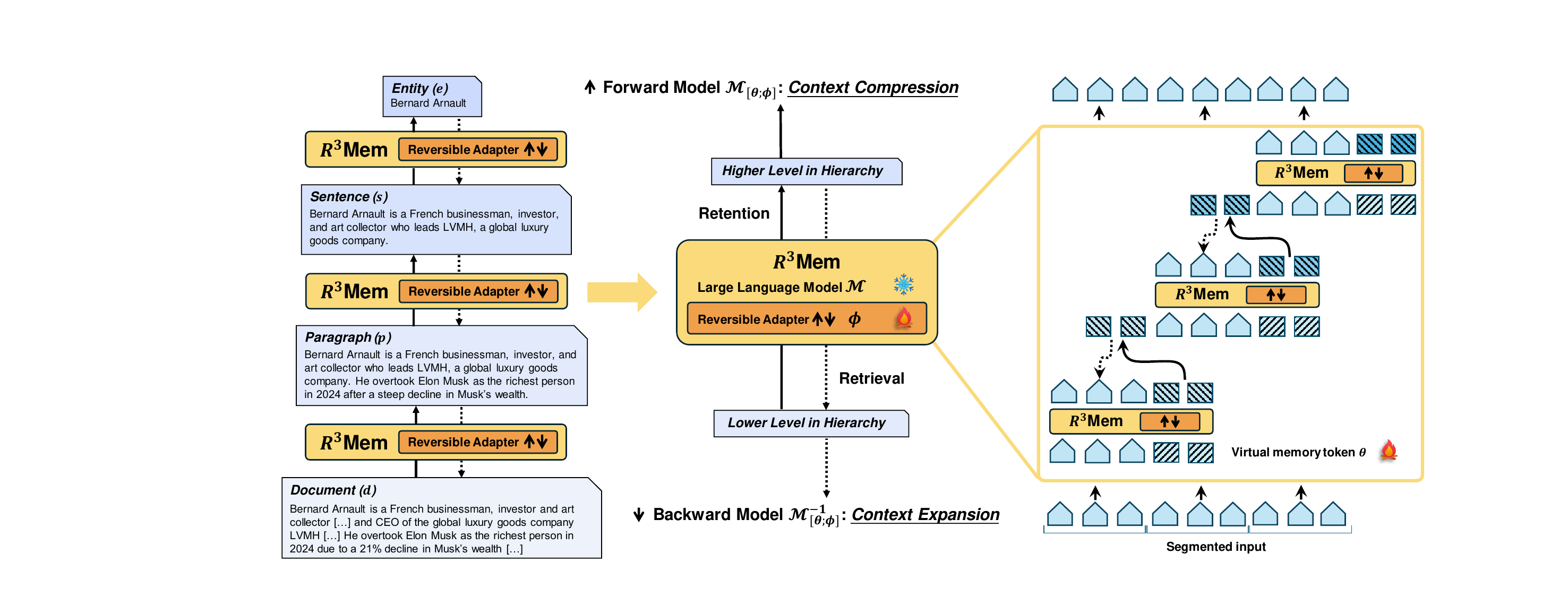}
    \caption{
    Overview of \ourmethod{}'s architecture: The model employs a reversible framework that integrates context compression and expansion mechanisms. 
    For the forward model, raw textual data is hierarchically encoded into compact representations at various levels—document, paragraph, and entity—using virtual memory tokens. In the backward model, the model reconstructs the original information by reversing the compression process.
    }
    \label{fig:pipeline}
    \vspace{-5mm}
\end{figure*}

In this section, we introduce \ourmethod{}, a memory network that optimizes both memory retention and retrieval.
As illustrated in Figure~\ref{fig:pipeline}, the core component of \ourmethod{} is \emph{context compression}, which encodes raw text into model parameters using \emph{virtual memory tokens}. These trainable tokens are appended to the raw text, summarizing the current context window and propagating information to subsequent windows. This enables the model to absorb and retain indefinitely long input sequences.
Furthermore, to facilitate more flexible memory usage, \ie, enabling retrieval of documents for queries with varying semantic granularities, we employ hierarchical compression. This approach chunks documents into multiple levels of semantic representation, including document-, paragraph-, sentence-, and entity-level abstractions. By structuring information hierarchically, our method optimizes retention and retrieval efficiency across different levels of granularity.
Lastly, we use a pre-trained Transformer backbone with an adapter-based reversible architecture, allowing the memory network to operate bidirectionally. This allows the model to be invoked in reverse, which reconstructs raw information from compressed memory akin to a ``zip'' and ``unzip'' process, unifying information retention and retrieval within a duplex network.

\subsection{Memory Retention}

Inspired by context-supervised pretraining~\citep{gao-callan-2022-unsupervised,w-etal-2023-query}, which trains models to generate one passage conditioned on another from the same document, we employ a similar mechanism to bridge the information gap between condensed memory and raw content. Specifically, we formulate memory retention as a context compression problem, where the model learns to generate a compressed representation (`query' $q$) given a raw text input (`context' $c$). 


To facilitate more flexible memory usage, we employ \textbf{hierarchical compression} to enhance multi-granularity assimilation, constructing $\langle c, q \rangle$ pairs at multiple levels, including \emph{document-to-paragraph}, \emph{paragraph-to-sentence}, and \emph{sentence-to-entity} mappings. This structured approach segments documents into different semantic granularities, ensuring optimized retention and adaptive retrieval across varying levels of abstraction.

Furthermore, we introduce \textbf{virtual memory tokens} to efficiently encode long contexts by splitting them into manageable segments and processing them sequentially while preserving previous information. These tokens cache and propagate memory across context windows, ensuring continuity in long-context retention and enabling the model to maintain coherence over extended sequences.

Formally, given a context-query pair $\langle c, q \rangle$ from the context-query set $D_c$, the memory network $\mathcal{M}_\theta$ learns to model
the conditional probability $\mathcal{M}_\theta (q \mid c )$ using an autoregressive decoder:
\begin{equation}
    \mathcal{M}_\theta (q \mid c ) = \prod_{t=1}^{T} \mathcal{M}_\theta (q_t \mid q_{<t}, c )
    \label{eq:forward_generation}
\end{equation}
where $T$ is the length of generated query comprised of a sequence of tokens $q = \langle q_1, \cdots, q_t, \cdots, q_T \rangle$, and $\theta$ denotes the virtual memory token.

\noindent
\textbf{Hierarchical compression.} \
To construct a structured hierarchy of text chunks, we borrow pipeline from \citet{xu2023recomp} and \citet{yoon2024compact} to employ a superior LLM to decompose each document $d$ into paragraphs $p$, sentences $s$, and key entities $e$ (detailed in Section~\ref{sec:experiments}). At each level, the preceding granularity (\eg, entire document) serves as the context and the subsequent (\eg, paragraphs) as the query, forming structured $\langle c, q \rangle$ pairs:
\begin{align}
    D_c &= D_d \cup D_p \cup D_s \\
        &= \left\{ \langle d, p \rangle \right\}_1^N \cup \left\{ \langle p, s \rangle \right\}_1^M \cup \left\{ \langle s, e \rangle \right\}_1^K
    \label{eq:hierarchical-compression}
\end{align}

\noindent
\textbf{Virtual memory tokens.} \
Encoding lengthy contexts, such as long-term interaction histories with LLMs, often exceeds the model's effective context window. As analyzed by \citet{an2024does}, even with a theoretically large context length, a model's ability to retain and effectively utilize relevant information remains limited in practice.
A naive approach would be to split long documents into smaller segments and process them individually. However, this disrupts semantic continuity and results in suboptimal training, as segmentation fragments contextual dependencies.

To address this, we introduce virtual memory tokens to bridge representations across context windows. These tokens act as summary vectors, caching compressed representations and transferring them across context windows.
Formally, as shown in Figure~\ref{fig:pipeline}, given a long input sequence $c$ and segmented as $c = c^1 \oplus c^2  \cdots \oplus c^s \cdots \oplus c^S$. we prepend and append memory tokens to each segment as:
$
    c^s = \inlinetcboxmathsuperscriptonly{$\theta$}{r} \oplus c^s \oplus \inlinetcboxmathsuperscriptonly{$\theta$}{w}
$
, where $\oplus$ denotes concatenation, \inlinetcboxmathsuperscriptonly{$\theta$}{r} represents the memory token outputs from the previous segment (with the $c^1$ having no such input), serving as memory \emph{read} tokens for the current segment, and \inlinetcboxmathsuperscriptonly{$\theta$}{w} represents the memory tokens of the current segment, acting as memory \emph{write} tokens to summarize the current segment and store information for future segments.
By leveraging virtual memory tokens, the model can scale beyond context length limitations while maintaining continuity across segments. 

Although token-based compression techniques have been widely explored in global attention~\citep{zaheer2020big,beltagy2020longformer}, our virtual memory tokens differ in that these tokens are trainable, enabling adaptive context compression and efficient optimization within a prompt-tuning paradigm~\citep{lester-etal-2021-power,liu2021p}. Moreover, they can further enhance memory capability by inserting memory tokens as hidden states within each Transformer layer~\citep{li-liang-2021-prefix}, as detailed in Section~\ref{subsec:in-depth-analysis}.

\begin{figure}[!t]
\centering
\includegraphics[width=0.95\columnwidth]{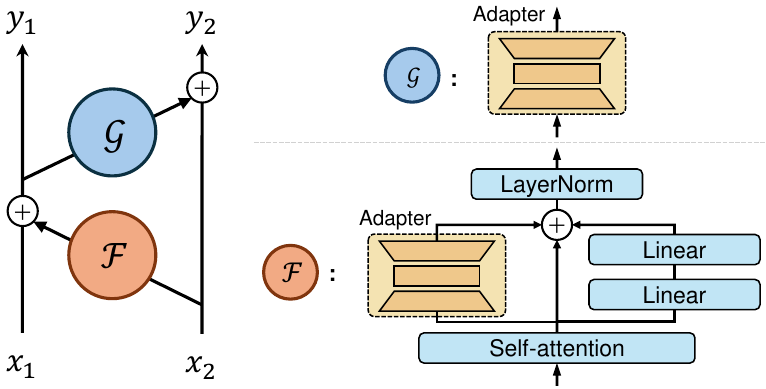}
\caption{The architecture of the reversible Transformer. Left: The general reversible neural architecture. Right: The components of the reversible Transformer.}
\label{fig:reversible-nn}
\vspace{-5mm}
\end{figure}

\subsection{Memory Retrieval}
Considering the dual nature of memory retention and retrieval, where retention integrates raw text into a compressed representation (\ie, compressing context into memory) and retrieval reverses this process by reconstructing the compressed representation into raw content (\ie, expanding memory to context), we propose building \ourmethod{} with a reversible architecture. By simply flipping the input and output ends, this approach enables a duplex transformation between context and its memory, allowing simultaneous optimization of memory retention and retrieval to improve retrieval accuracy.

As shown in Figure~\ref{fig:reversible-nn}, reversible architectures are a class of neural networks based on NICE~\citep{dinh2014nice,dinh2022density}, which construct nonlinear bijective transformations by partitioning input at each layer into two groups that cache information for one another, thereby allowing exact reconstruction of inputs. 
Since the standard Transformer architecture is not inherently reversible, \citet{liao2024make} introduced adapter-based modifications to pre-trained Transformers to make them reversible. The key idea is to treat the original Transformer layer as one input group and the inserted adapter module as another, forming a reversible Transformer where the adapters are optimized using adapter tuning~\citep{houlsby2019parameter,hu2021lora}. We provide a more detailed introduction of reversible Transformer in Appendix~\ref{appx:reversible_transformers}.

We use the pre-trained Transformer-based LLaMA 3.1-8B as the base model and integrate adapter modules to enable a reversible architecture. 
This allows us to reconstruct the input by feeding the compressed content backward.
Formally, we denote the flipped model as $\mathcal{M}^{-1}$, where the backward generation models a similar conditional probability as the forward process in Eq.~\ref{eq:forward_generation}:
\begin{equation}
    \mathcal{M}_{[\theta;\phi]}^{-1} (c \mid q ) = \prod_{l=1}^{L} \mathcal{M}_{[\theta;\phi]}^{-1} (c_l \mid c_{<l}, q)
    \label{eq:backward_generation}
\end{equation}
where $L$ is the length of the generated context, encompassing a sequence of tokens $c = \langle c_1, \cdots, c_l, \cdots, c_L \rangle$, and $\phi$ is the adapter matrix.

\subsection{Training Objective}

Following the standard training setup of reversible architectures~\citep{he2016dual,zheng2021duplex,wu-2023-duplex}, we optimize \ourmethod{} through bidirectional training with cycle consistency, incorporating forward compression loss, backward expansion loss, and a cycle consistency loss.
\begin{align}
    \mathcal{L} = \mathcal{L}_{\text{forward}} + \mathcal{L}_{\text{backward}} + \lambda \mathcal{L}_{\text{cycle}}
    \label{eq:final_loss}
\end{align}
where $\lambda$ is the coefficient to balance the contribution of cycle consistency loss.

Given a context-query pair $\langle c, q\rangle \in D_c$, forward training optimizes the memory network to model the probabilities of forward generation, as defined in Eq.~\ref{eq:forward_generation}, by minimizing the conditional negative log-likelihood (NLL) loss:
\begin{align}
    \mathcal{L}_{\text{forward}} = -\sum_{t=1}^{T}{\log \widehat{\mathcal{M}}_{[\theta;\phi]} \left(q_t \mid q_{<t}, c \right)}
\end{align}
where $\widehat{\mathcal{M}}_{[\theta;\phi]} \left(q_t \mid q_{<t}, c \right)$ represents the predicted probability for token $q_t$ in the reference query.

Similarly, backward training models the probabilities of backward generation as defined in Eq.~\ref{eq:backward_generation}:
\begin{align}
    \mathcal{L}_{\text{backward}} = -\sum_{l=1}^{L}{\log \widehat{\mathcal{M}}_{[\theta;\phi]}^{-1} \left(c_l \mid c_{<l}, q \right)}
\end{align}
where $\widehat{\mathcal{M}}_{[\theta;\phi]}^{-1} \left(c_l \mid c_{<l}, q \right)$ denotes the predicted probability for $c_l$ in the reconstructed context.

To ensure cycle consistency, given an input $c$, we generate its reconstruction $\bar{c}$ by passing it through the reversible model as the forward mapping $f_{\mathcal{M}}$ and backward mapping $f_{\mathcal{M}^{-1}}$:
\begin{align}
    f_{\mathcal{M}} &: c \longmapsto q \\
    f_{\mathcal{M}^{-1}} &: q \longmapsto \bar{c}
\end{align}
The cycle consistency loss maximizes the similarity between the original input $c$ and its reconstruction $\bar{c}$ using cross-entropy:
\begin{align}
     \mathcal{L}_{\text{cycle}} = -\sum_{l=1}^{L}{\log \widehat{\mathcal{M}}_{[\theta;\phi]}^{-1} \left(c_l \mid c_{<l}, f_{\mathcal{M}} (c) \right)}
\end{align}

\section{Experiments}
\label{sec:experiments}

\noindent
\textbf{Dataset.} \
We follow the training protocol of \citet{qian2024memorag} using UltraDomain~\citep{qian2024memorag},
which includes documents from the training set of diverse long-context question-answering and summarization tasks, including NarrativeQA~\citep{kocisky-etal-2018-narrativeqa}, Qasper~\citep{dasigi-etal-2021-dataset}, GovReport~\citep{huang-etal-2021-efficient}, and MultiNews~\citep{fabbri-etal-2019-multi}.
Following~\citet{xu2023recomp,yoon2024compact}, we employ a more capable LLM as an oracle to generate hierarchical context-query pairs.
While certain lightweight doc2query models~\citep{nogueira2019document,w-etal-2023-query} demonstrate strong performance in constructing context-query pairs, they often struggle with longer inputs. In Section~\ref{subsec:in-depth-analysis}, we present an experimental comparison between context-query pairs generated by doc2query models and the oracle model.

Specifically, we prompt a high-capacity oracle model (\ie, GPT-4o) to progressively decompose each document into paragraphs, sentences, and entities. This process draws inspiration from event-centric hierarchical summarization methods~\citep{zhong-etal-2022-unsupervised,zhu2024eventsum}. Firstly, given a document $d$, the oracle generates a set of query-worthy events and selects the most relevant entities. 
We then prompt the oracle to retrieve sentence-level contexts, \ie, $s$, surrounding these entities and condense them into sentence-entity pairs $\langle s, e \rangle$. 
Building on these pairs, we instruct the oracle to extend and summarize the retrieved context into paragraph-level chunks, \ie, $p$, to create document-paragraph pairs $\langle d, p \rangle$ and paragraph-sentence pairs $\langle p, s \rangle$.
Finally, we apply a length-based criterion to filter out short paragraphs and sentences. Paragraphs shorter than 20\% of the corresponding original document length and sentences shorter than 4\% are removed.
The statistics of these constructed pairs are presented in Table~\ref{tab:context-query-statistics} and the used prompts are provided in Appendix~\ref{sec:experiment-setup}.

\noindent
\textbf{Baselines.} \
We compare \ourmethod{} with five memory-augmented Transformer architectures, categorized into recurrent architectures and associative memory architectures. The former include RMT~\citep{bulatov2022recurrent}, MemoRAG~\citep{qian2024memorag}, and MELODI~\citep{chen2024melodi}, while the latter comprise MemoryLLM~\citep{wang2024memoryllm} and CAMELoT~\citep{he2024camelot}. 
For RMT, MemoryLLM, and MemoRAG, we utilize their official implementations to report results. Since MELODI and CAMELoT have not publicly released their code, we report their results as presented in their respective papers and ensure that our evaluation settings align with theirs for a fair comparison.
Implementation details are provided in Appendix~\ref{sec:implementation-details}.

\subsection{Retention Performance}
\label{subsec:compression-performance}
We firstly demonstrate whether \ourmethod{} can effectively compress and encode context.
Following the setting of MELODI~\citep{chen2024melodi}, we assess retention performance by measuring perplexity in long-context language modeling across three publicly available datasets: PG19~\citep{rae2019compressive} , Pile arXiv~\citep{gao2020pile}, and C4 (4K+)~\citep{raffel2020exploring}.
The detailed experimental setup is provided in Appendix~\ref{sec:experiment-setup}.
The average perplexity on the testing set is summarized in Table~\ref{tab:compression-result-1}.


\noindent
\textbf{\ourmethod{} achieves state-of-the-art performance in long-context modeling}.\
\ourmethod{} attains the lowest perplexity across all three datasets, effectively compressing long contexts into memory vectors. Notably, on the challenging C4 (4K+) dataset, \ourmethod{} reduces the perplexity by approximately 13\% compared to the next-best baseline, MemoRAG.

\noindent
\textbf{Recurrent architectures outperform associative memory}.\
For example, associative memory-based methods, such as MemoryLLM and CAMELoT, exhibit inferior performance compared to the others. This disparity may stem from the reliance of associative memory approaches on discrete read and write operations, as seen in CAMELoT, and drop operations, as in MemoryLLM, which may struggle to maintain smoothly evolving contextual representations over long sequences.

\begin{table}[!t]
    \centering
     \resizebox{0.8\columnwidth}{!}{
        \begin{tabular}{c|ccc}
            \toprule
            \textbf{Model}& \textbf{PG19}& \textbf{arXiv}& \textbf{C4 (4K+)} \\
            \midrule
            \midrule
            MemoryLLM& 7.65& 4.00& 18.14 \\
            CAMELoT& 7.10& 3.60& - \\
            \hline
            RMT& 7.04& 3.56& 17.67 \\
            MELODI& 6.21& -& 15.25 \\
            MemoRAG& 5.92& 3.35& 15.37 \\
            \hline
            \textbf{\ourmethod{}}& \textbf{5.21}& \textbf{2.39}& \textbf{13.38} \\
            \bottomrule
        \end{tabular}
     }
    \caption{Long-context language modeling performance in terms of perplexity among three benchmarks.
    The dash ``-'' indicates that the code is not publicly available and corresponding results are not reported in their paper.
    }
    \label{tab:compression-result-1}
    \vspace{-5mm}
\end{table}

\subsection{Retrieval Performance}
\label{subsec:retrieval-performance}
We further validate that the encoded information can be faithfully retrieved, establishing a reliable foundation for retrieval tasks. To assess this, we follow the experimental setup of MemoRAG~\citep{qian2024memorag} and integrate \ourmethod{} into a retrieval-augmented generation (RAG) question-answering (QA) task on UltraDomain, using the same in-domain and out-of-domain evaluation settings.

\begin{figure}[!t]
\centering
    \includegraphics[width=0.8\columnwidth]{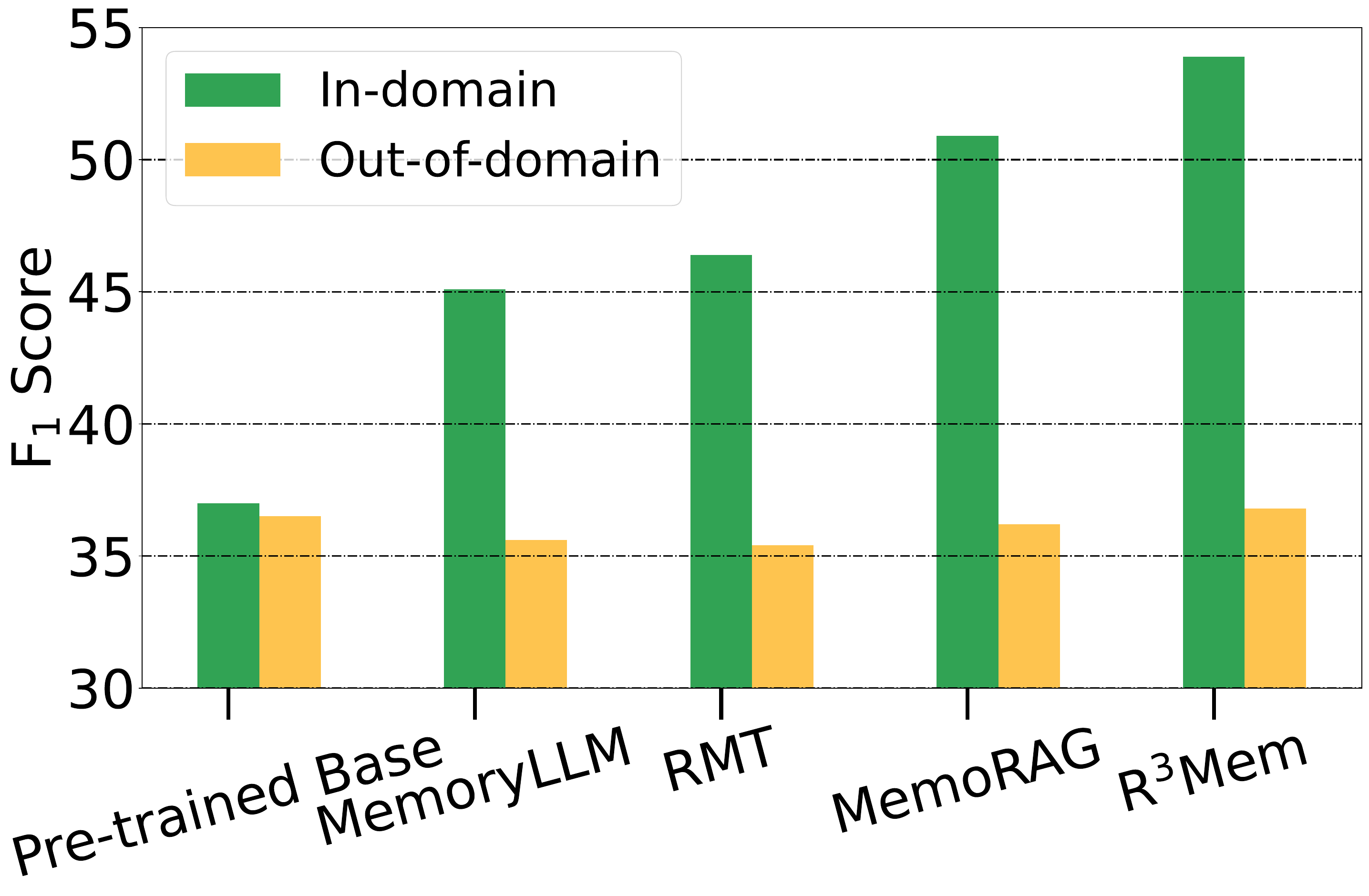}
    \caption{RAG performance on the UltraDomain dataset in terms of in-domain and out-of-domain settings.}
    \label{fig:model-rag-performance}
    \vspace{-5mm}
\end{figure}

During evaluation, the model receives only the query as input, without direct access to the original test set context. This setup allows us to assess whether the model can effectively recall and utilize encoded context to generate accurate responses. The average F$_1$ Scores are shown in Figure~\ref{fig:model-rag-performance}.
The results indicate that \textbf{retrieval performance is closely aligned with compression performance}, \ie, better compression leads to improved retrieval.
This demonstrates the dual nature of context compression and expansion, highlighting the rationale behind \ourmethod{} optimizing both to enhance retrieval accuracy.
Notably, \ourmethod{} achieves the best results in both retrieval and retention performance, while MemoryLLM underperforms in both aspects.

\textbf{Out-of-domain performance is significantly lower than in-domain performance, but remains consistent across models}.
The out-of-domain performance is notably lower than in-domain performance, as observed in \ourmethod{}'s results (in-domain: 53 vs. out-of-domain: 36). In contrast, the base pre-trained model exhibits a smaller gap (in-domain: 38 vs. out-of-domain: 37), suggesting that fine-tuning on domain-specific data has minimal impact on out-of-domain generalization. This indicates that integrating new context does not degrade the model's original knowledge.
Additionally, the performance differences across models in the out-of-domain evaluation are relatively small. \ourmethod{} achieves only a marginal 1\% improvement over the weakest-performing MemoryLLM, and all baselines exhibit out-of-domain performance similar to the base model. To further investigate this phenomenon, we scale training iterations and analyze its effects in Section~\ref{subsec:in-depth-analysis}.

\subsection{Agent Performance}
We assess \ourmethod{} in a real-world agent data using SiliconFriend~\citep{zhong2024memorybank}, an AI chatbot companion. Specifically, we replace its external memory module, \ie, MemoryBank~\citep{zhong2024memorybank} with \ourmethod{}. 
We use the publicly available SiliconFriend dataset, which consists of interactions among 15 distinct virtual users over a 10-day period. Following the setup of \citet{zhong2024memorybank}, where the external memory bank is initialized with given dialogue history, we initialize the implicit memory module by training \ourmethod{} on this dialogue history for two epochs. For comparison, we employ MemoRAG (\ie, the most competitive baseline in Table~\ref{tab:compression-result-1}) as the baseline implicit memory module and fine-tune it under the same settings.

We use 194 memory-probing questions. First, the models retrieve context from the memory bank or generate context using MemoRAG and \ourmethod{}. The retrieved or generated context is then fed into SiliconFriend to generate final responses. 
The evaluation includes four key metrics: (1) \emph{Memory Retrieval Accuracy}, measuring alignment with the reference memory using the F$_1$ Score; (2) \emph{Response Correctness}, assessing whether the response contains the correct answer via exact substring matching; (3) \emph{Contextual Coherence}, evaluating response fluency and relevance using BARTScore-Faithfulness~\citep{yuan2021bartscore}; and (4) \emph{Ranking Score}, ranking memory modules based on response correctness, with scores computed as $s = 1/r$, where $r\in\{1,2,3\}$ denotes ranking position. 

\begin{figure}[!t]
\centering
    \includegraphics[width=0.8\columnwidth]{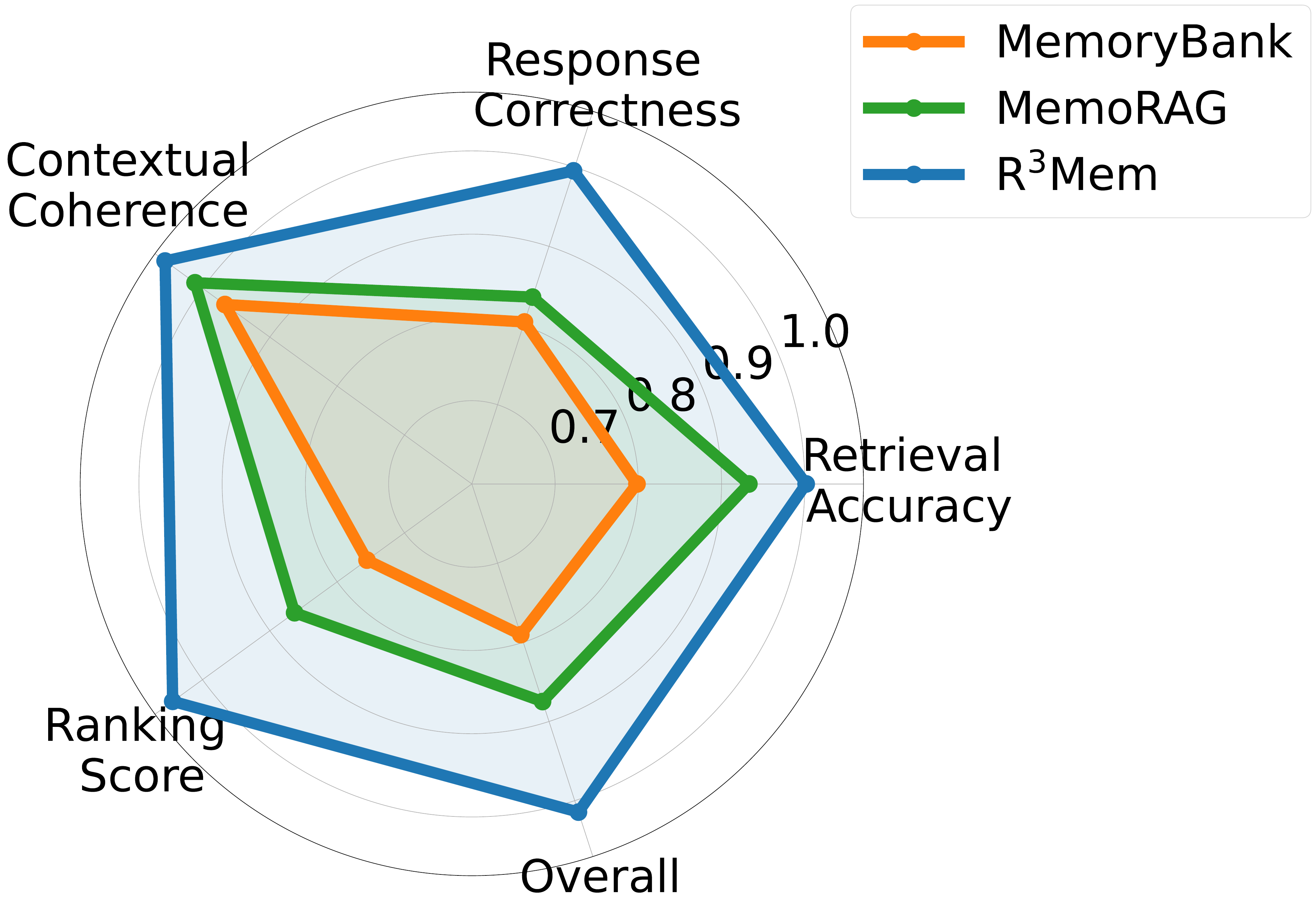}
    \caption{Evaluation of memory retrieval and response generation when integrating \ourmethod{} into the SiliconFriend conversational agent.
    The overall score represents the average across all four evaluation metrics.
    Scores are re-scaled using min-max normalization for each metric to enhance clarity.
    }
    \label{fig:model-agent-performance}
    \vspace{-5mm}
\end{figure}

The results, summarized in Figure~\ref{fig:model-agent-performance}, reveal two key findings. Firstly, \textbf{implicit memory modules outperform explicit memory}. For instance, both \ourmethod{} and MemoRAG surpass the original MemoryBank across all four metrics. Beyond \emph{Ranking Scores} and \emph{Response Correctness}, \emph{Memory Retrieval Accuracy} exhibits the most significant difference between implicit memory modules and explicit memory. Notably, \ourmethod{} achieves the best overall performance, primarily due to its superior \emph{Memory Retrieval Accuracy}.
However, \textbf{Contextual Coherence shows no significant differences across memory modules}. This could be due to the fact that SiliconFriend has been fine-tuned on psychological dialogues, enabling it to generate fluent and natural responses even when the retrieved memory is not entirely accurate.

\subsection{In-depth Analysis}
\label{subsec:in-depth-analysis}
\noindent
\textbf{Hierarchical compression and high-quality context-query pairs improve performance.} \
We construct two baseline models that exclude hierarchical context-query pairs: (1) \textbf{\ourmethod{}-context-only}, which retains only document-paragraph pairs (\ie, $\langle d, p \rangle$ in Eq.~\ref{eq:hierarchical-compression}), and (2) \textbf{\ourmethod{}-short-context}, which generates context-query pairs using a lightweight doc2query model~\citep{w-etal-2023-query}. The latter feeds document chunks of fewer than 512 tokens as input and generates queries.

We train both baselines using same setting as \ourmethod{}. As shown in Table~\ref{tab:ablation-results}, removing hierarchical compression or restricting context-query length significantly increases perplexity. The short-context variant leads to an even larger performance drop, reducing retention effectiveness by approximately 28\% on C4 (4K+). These results highlight the necessity of hierarchical compression for effectively encoding context of \ourmethod{}.

\begin{table}[!t]
    \centering
     \resizebox{0.9\columnwidth}{!}{
        \begin{tabular}{c|ccc}
            \toprule
            \textbf{Model}& \textbf{PG19}& \textbf{arXiv}& \textbf{C4 (4K+)} \\
            \midrule
            \midrule
            Fine-tuned Base& 8.10& 4.31& 19.04 \\
            \hline
            \ourmethod{}-context-only& 7.15& 3.69& 17.16 \\
            \ourmethod{}-short-context& 7.44& 3.83& 17.60 \\
            \hline
            \ourmethod{}-w/o-backward& 6.41& 3.21& 15.83 \\
            \ourmethod{}-w/o-cycle& 5.91& 2.87& 14.80 \\
            \hline
            \textbf{\ourmethod{}}& \textbf{5.21}& \textbf{2.39}& \textbf{13.38} \\
            \bottomrule
        \end{tabular}
     }
    \caption{Ablation analysis of \ourmethod{} on long-context language modeling task.}
    \label{tab:ablation-results}
    \vspace{-2mm}
\end{table}

\begin{figure}[!t]
    \centering
	\begin{subfigure}{0.48\columnwidth}
        \includegraphics[width=\columnwidth]{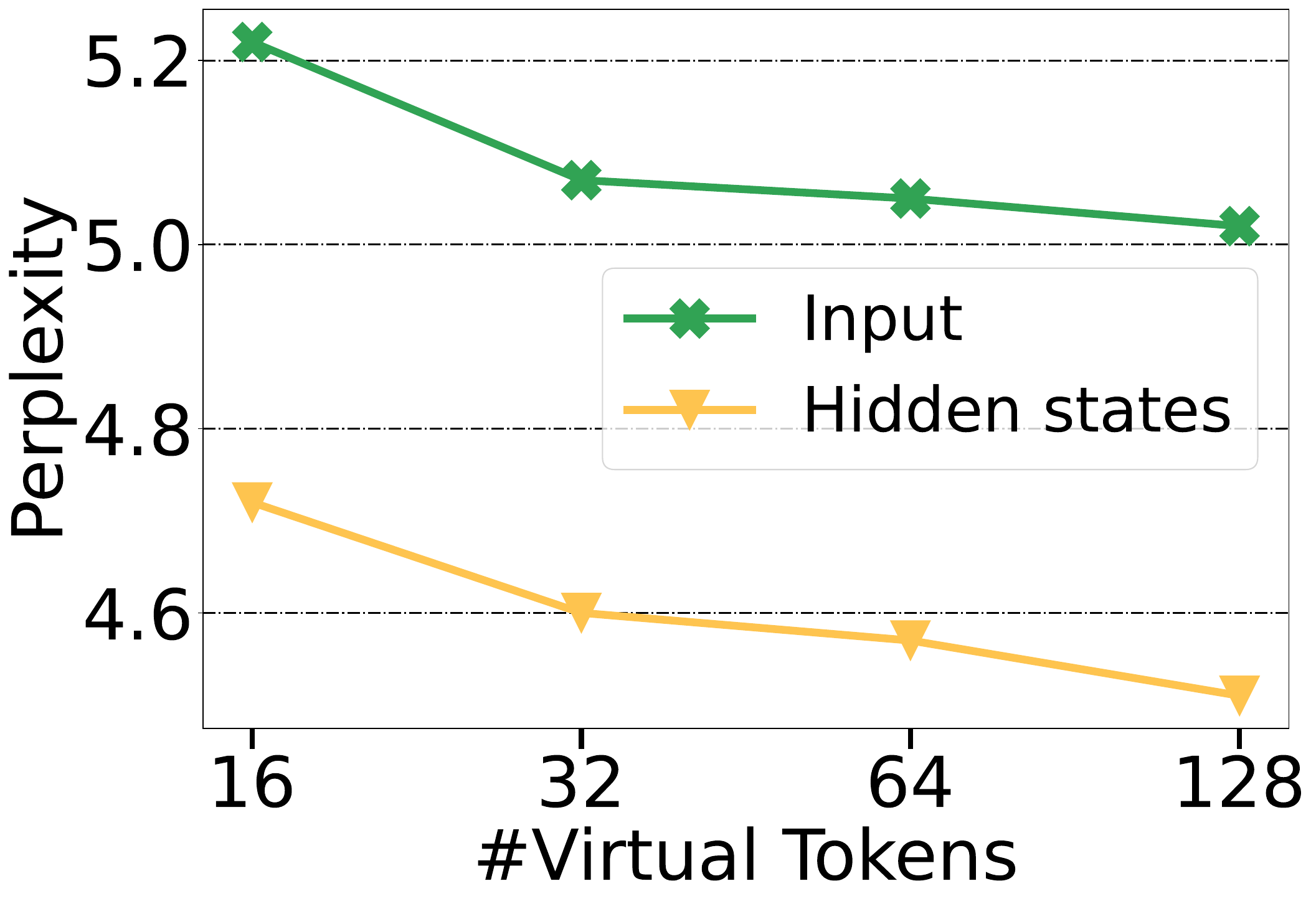}
	\end{subfigure}
	\begin{subfigure}{0.49\columnwidth}
        \includegraphics[width=\columnwidth]{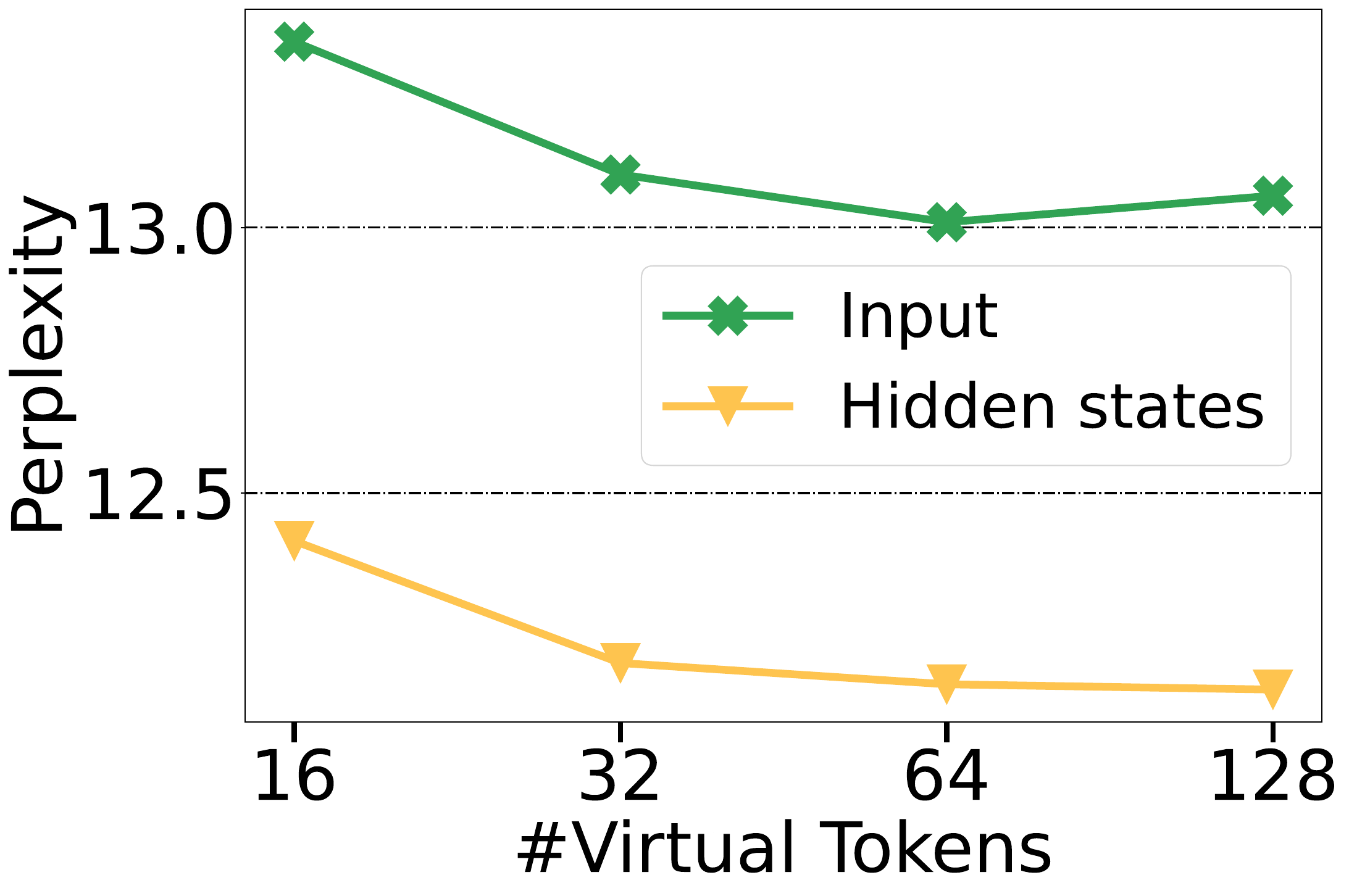}
    \end{subfigure}
    \caption{Long-context language modeling performance on the PG19 (left) and C4 (4K+) (right) dataset when scaling virtual memory tokens with increasing number and injection of hidden states.}
    \label{fig:vitural-token-scaling}
    \vspace{-5mm}
\end{figure}

\noindent
\textbf{Injecting virtual tokens into hidden states improves performance, but increasing token length does not.} \
We assess the impact of scaling virtual memory tokens: (1) increasing the number of input virtual tokens from 16 to 32, 64, and 128, and (2) injecting virtual tokens into the hidden states of each Transformer layer, following \citet{li-liang-2021-prefix}, with scaling from 16 to 32, 64, and 128.

The results, illustrated in Figure~\ref{fig:vitural-token-scaling}, reveal two key observations. First, adding virtual tokens to hidden states improves final performance, suggesting that similar to hierarchical compression, hierarchical memory across Transformer layers enhances memory retention. However, this approach drastically increases the number of trainable parameters by a factor of 32 in the LLaMA 3.1-8B base model, making it impractical for large-scale deployment. As a result, \ourmethod{} defaults to using virtual tokens only in the input sequence. Second, increasing token length (\ie, memory size) does not significantly improve performance, which is inconsistent with prompt-tuning. 
A possible explanation is that memory tokens serve a different function from prompt tuning. While prompt tuning enhances task-level generalization by stimulating pre-trained knowledge, memory tokens summarize and store sample-level context, making them easier to fit than multi-task learning.

\begin{figure}[!t]
\centering
    \includegraphics[width=0.8\columnwidth]{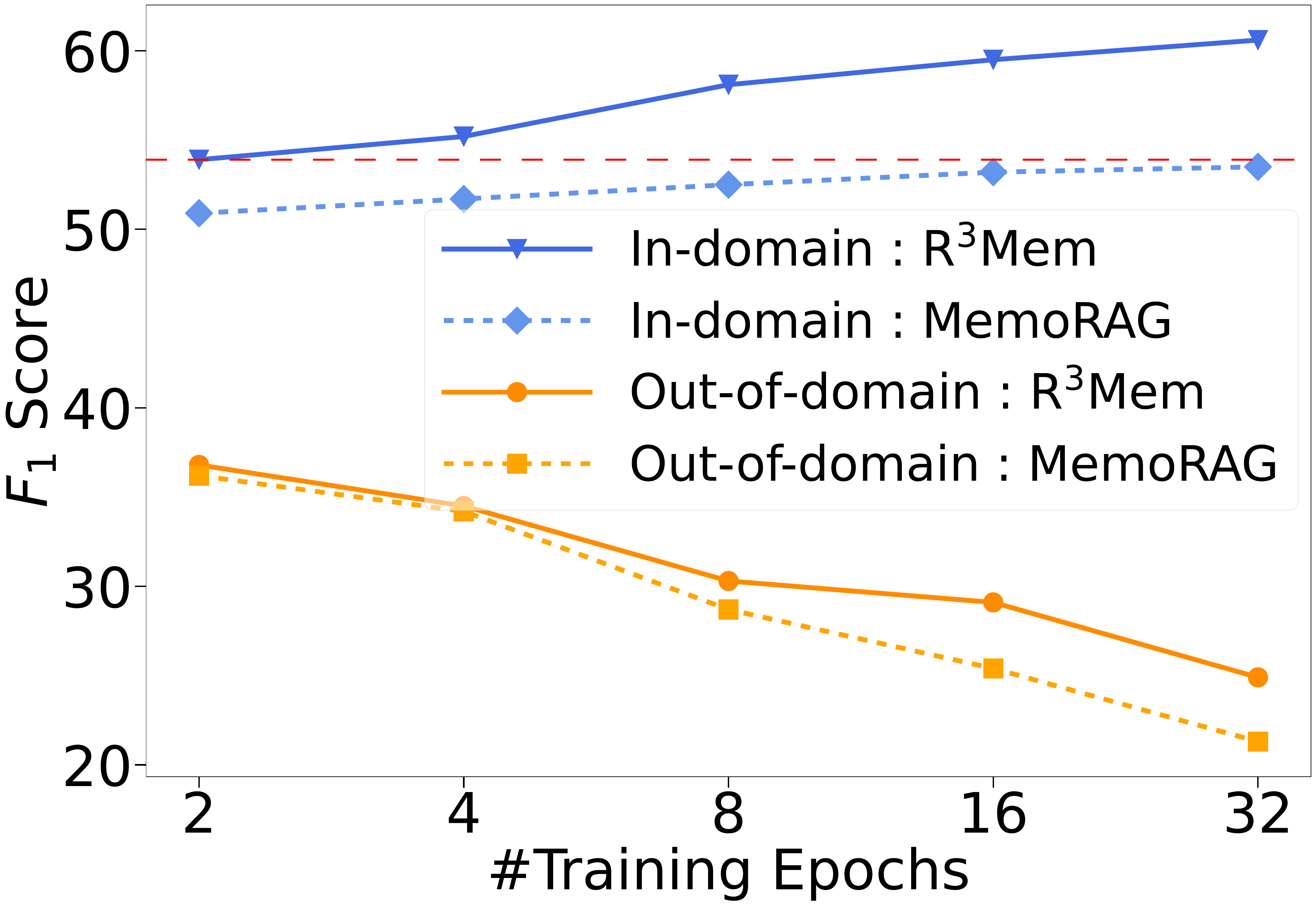}
    \caption{RAG performance when strengthening memory through training extra epochs on the training set.}
    \label{fig:training-iteration}
    \vspace{-5mm}
\end{figure}

\noindent
\textbf{Backward optimization and cycle consistency loss are essential.} \
We investigate the effect of training objectives by testing two ablated variants: (1) \textbf{\ourmethod{}-w/o-backward}, which removes the backward loss, and (2) \textbf{\ourmethod{}-w/o-cycle}, which omits cycle consistency loss. As shown in Table~\ref{tab:ablation-results}, both variants exhibit degraded performance, demonstrating that all three loss components contribute to optimal training by ensuring alignment between context compression and expansion. This alignment is key to \ourmethod{}, enabling memory retention and retrieval in a duplex framework.

\noindent
\textbf{Extra training improves in-domain retrieval at the cost of out-of-domain generalization.} \
We analyze whether overfitting to the training set strengthens memorization of training contexts. We extend training from 2 epochs to 4, 8, 16 and 32 epochs and evaluate in-domain and out-of-domain retrieval performance. Results in Figure~\ref{fig:training-iteration} indicate that while in-domain retrieval consistently improves with prolonged training, it comes at the cost of out-of-domain generalization, observed in both \ourmethod{} and MemoRAG.
This suggests that although implicit memory avoids the management and storage overhead of explicit memory, integrating new memory through fine-tuning may be unstable, as newly encoded memory may interfere with or even overwrite pre-trained parametric knowledge, making it harder for effective lifelong integration.


\section{Related Works}
\label{sec:related works}

\noindent \textbf{Memory-augmented neural networks.}\ 
Designing architectures capable of memorization and generalization through knowledge abstraction~\citep{sukhbaatar2019augmenting} and data-dependent information retention~\citep{zancato2024b} has been a longstanding research focus. 
Early approaches introduced architectures with external memory modules, such as neural turing machines (NTM)~\citep{graves2014neural} and modern Hopfield Networks~\citep{ramsauer2020hopfield}, which utilize pre-defined update rules to manage memory. With the advent of Transformers, some methods employ recurrent Transformer architectures~\citep{dai2019transformer,bulatov2022recurrent} to cache key-value pairs as memory, enabling the reuse of cached information to extend context window sizes. 
Additionally, recent studies have explored encoding training data into model parameters, effectively using them as memory to store world knowledge~\citep{wang2024self,padmanabhan2024propagating,gangadhar-stratos-2024-model,he2024mixture}. This approach has also been extended to large databases~\citep{qian2024memorag}, test-time data points~\citep{sun2024learning}, and broader language modeling tasks~\citep{yang2024text}.
Titans~\citep{behrouz2024titans} integrates long-term, short-term, and persistent memory into a unified neural architecture.
While optimizing memory retention, they overlook retrieval reliability from model parameters, which is a core design motivation of \ourmethod{}.

\noindent \textbf{Context compression.}\ 
Compressing lengthy contexts into concise representations that retains essential information can make LLM inference more efficient~\citep{choi2022prompt,li2024prompt}.
Approaches like Selective Context~\citep{li-etal-2023-compressing}, LLMLingua~\citep{jiang-etal-2023-llmlingua,jiang-etal-2024-longllmlingua} and RECOMP~\citep{xu2023recomp} use context selection to improve inference efficiency, and methods such as AutoCompressor~\citep{chevalier-etal-2023-adapting}, in-context autoencoder (ICAE)~\citep{ge2023context}, Gist~\citep{mu2024learning} and CompAct~\citep{yoon-etal-2024-compact} employ training-based techniques to generate summary representations. 
Besides, \citet{deletang2023language} proposes new general-purpose language modeling perspectives by leveraging compression through arithmetic coding from information theory~\citep{rissanen1976generalized,pasco1976source}. 
In contrast, \ourmethod{} uses context compression as a surrogate task to optimize memory retention while ensuring alignment through backward context expansion.

\section{Conclusion}
\label{sec:conclusion}
We propose \ourmethod{}, a memory network built on a reversible architecture that optimizes both information retention and retrieval. \ourmethod{} employs hierarchical compression to adaptively process input and utilizes virtual memory tokens to encode long-context information. Empirical results demonstrate state-of-the-art performance in long-context modeling and retrieval, with strong scalability and accuracy in real-world conversational agents.

\section*{Limitations}
\label{sec:limitations}

\ourmethod{} is a duplex network that unifies memory retention and retrieval, which learns to encode documents into model parameters through context compression. The limitations of this framework fall into two main aspects: (1) the trade-off between high-quality memory retention and the complexity and cost of the context-query construction pipeline, and (2) the instability of implicit memory in lifelong integration.

On the one hand, ensuring effective document assimilation requires hierarchical compression optimized through hierarchical context-query pairs. As analyzed in Section~\ref{subsec:in-depth-analysis}, the quality of these pairs significantly impacts memory retention effectiveness. For example, while a lightweight doc2query model produces reasonably good results, they are still less effective than those generated by more capable LLMs, which, in turn, come with significantly higher computational costs. Balancing high-quality memory retention with the complexity and cost of the context-query construction pipeline is crucial. Depending on the frequency of new context integration into the model parameters, incorporating an adaptive data construction pipeline within \ourmethod{} could enhance its efficiency, making this an important direction for future work.

On the other hand, strengthening memory retention through additional training may impact pre-trained parametric knowledge or overwrite existing memory (as analyzed in Section~\ref{subsec:in-depth-analysis}). While this behavior may be desirable in narrow-domain applications or when maintaining a small-scale memory history, excessive training could undermine the model's inherent contextual understanding and commonsense reasoning. Further evaluation is needed to comprehensively understand these relationships. Additionally, developing a more controllable memory architecture that better balances historical context retention with new knowledge integration, such as incorporating expert network routing mechanisms, remains an important avenue for future work on \ourmethod{}.

\section*{Ethics Statement}
\ourmethod{}, as a memory network, can be integrated into memory-intensive applications~\citep{zhang2024survey} such as social simulation, conversational assistants, and personalized recommendations. While the model can encode personal information from interaction history of applications into its parameters, its training-based nature allows for the filtering of harmful or sensitive content during memory construction. This ensures a personalized experience and optimal retrieval performance while safeguarding users from potential harm.

\bibliography{allinone}

\clearpage
\appendix

\section{Implementation Details}
\label{sec:implementation-details}
We developed our method using PyTorch~\citep{paszke2019pytorch}. We initialize the base model, LLaMA 3.1-8B, with checkpoints from the Hugging Face Transformers package~\citep{wolf-etal-2020-transformers}. 
We implemented the adapter module using LoRA~\citep{hu2021lora}, setting the scaling factor $\alpha=32$ and the rank $r=8$, with a dropout of $0.1$ applied. 
The default setting for the number of virtual memory tokens is $8$, unless scaled to $16$, $32$, $64$, or $128$, as discussed in Section~\ref{subsec:in-depth-analysis}. These tokens are randomly initialized by sampling from $\mathcal{N}(0, 0.02)$. We set the coefficient of the cycle consistency loss $\lambda$ in Eq.~\ref{eq:final_loss} to $0.5$.
Fine-tuning is performed for $2$ epochs using the AdamW optimizer~\citep{loshchilov2018decoupled} with a maximum learning rate of $2\times 10^{-5}$, $\beta_1=0.9$, $\beta_2=0.99$, and a learning rate warmup period covering 6\% of the total training steps. 
In Section~\ref{subsec:in-depth-analysis}, we scale the training epochs to $4$, $8$, $16$, and $32$. 
The batch size is $2$, and training is conducted on a single NVIDIA RTX A5000 24 GB GPU, taking approximately $13$ hours for a 2-epoch run. 
Experiments were conducted over four independent runs with different random seeds, and the best evaluation results were selected for reporting.
For RMT, MemoryLLM, and MemoRAG, we utilize their official implementations to report results. For MELODI and CAMELoT, as their code is not publicly available, we report their results as stated in their respective papers and ensure that our evaluation settings align with theirs for a fair comparison.


\section{Experiment Setup}
\label{sec:experiment-setup}

\noindent
\textbf{Context-query pairs.}
The prompt used to construct the hierarchical context-query pairs is presented in Figure~\ref{fig:prompt}. We use a low temperature of 0.3 and greedy decoding to preserve accurate event-related details. The statistics of the constructed context-query pairs are summarized in Table~\ref{tab:context-query-statistics}.

\noindent
\textbf{Long context language modeling.} \
We firstly demonstrate whether \ourmethod{} can effectively compress and encode context.
Specifically, following the setting of MELODI~\citep{chen2024melodi}, we assess compression performance by measuring perplexity in long-context language modeling across three publicly available datasets: PG19~\citep{rae2019compressive} , arXiv from the Pile~\citep{gao2020pile}, and C4 (4K+)~\citep{raffel2020exploring}.

While MELODI also evaluates on a custom-collected dataset from arXiv Bulk Data Access, details about their data pipeline and cleaned data are not publicly available. Instead, we utilize the arXiv subset from the Pile dataset~\citep{gao2020pile}, which comprises technical papers in mathematics, computer science, and physics, totaling 1,264 documents in the test split.

The PG19 test set includes 100 English books, each containing 68,972 tokens on average. For the C4 dataset, a web-crawled corpus of internet documents, we employ the ``c4/en'' subset, which has undergone cleaning and deduplication. To focus on long-context scenarios, we filter out samples with fewer than 4,096 tokens, resulting in 155,007 testing samples.

\noindent
\textbf{Retrieval-augmented generation.} \
We further validate whether the encoded memory can be faithfully retrieved, establishing a reliable foundation for retrieval tasks. To assess this, we follow the experimental setup of MemoRAG~\citep{qian2024memorag} and integrate \ourmethod{} into a retrieval-augmented generation (RAG) question-answering (QA) task on UltraDomain, using the same in-domain and out-of-domain evaluation settings.

For in-domain evaluation, we use a subset of the UltraDomain test set, where both the training and test samples are based on the same underlying world knowledge. This knowledge is sourced from Wikipedia, research papers in S2ORC~\citep{lo-etal-2020-s2orc}, ebooks from Project Gutenberg\footnote{\url{https://www.gutenberg.org/}}, and domain-specific financial and legal documents. 
For out-of-domain evaluation, we use another subset of the UltraDomain test set, where queries and contexts are drawn from textbooks spanning 18 diverse domains, including biology, religion, art, etc. This evaluation measures the model's ability to retrieve and apply knowledge that was not explicitly present in the training data, testing its generalization beyond the training distribution.

\begin{table}[!t]
    \centering
    \resizebox{1.0\columnwidth}{!}{
        \begin{tabular}{ccccc}
            \toprule
            & \textbf{Document}& \textbf{Paragraph}& \textbf{Sentence}& \textbf{Entity} \\
            \midrule
            \midrule
            \#Samples& 2,178& 10,198& 50,989& 152,968 \\
            Max. Length& 9,528& 1,803& 42& 12 \\
            Min. Length& 1,356& 207& 364& 1 \\
            Avg. Length& 7,470& 1,537& 319& 3 \\
            \bottomrule
        \end{tabular}
    }
    \caption{Breakdown of statistics for the constructed context-query pairs. Length indicates the length of each text chunk, measured in tokens.
    }
    \label{tab:context-query-statistics}
    \vspace{-5mm}
\end{table}

\noindent
\textbf{Conversational agent.} \
We use 194 memory-probing questions. First, the models retrieve context from the memory bank or generate context using MemoRAG and \ourmethod{}. The retrieved or generated context is then fed into SiliconFriend to generate final responses. 

The evaluation covers four key metrics: \textbf{(1) Memory retrieval accuracy}: It measures the alignment of retrieved memory with reference memory using the F$_1$-score.
\textbf{(2) Response correctness}: It evaluates whether the response contains the correct answer. Since the gold answer may be embedded within a longer dialogue response, correctness is determined using exact substring matching. \textbf{(3) Contextual coherence}: It assesses whether the response is natural and coherent within the given context and dialogue history. This is evaluated using BARTScore-Faithfulness~\citep{yuan2021bartscore}, a widely used automatic metric for natural language generation that measures the relevance of the candidate response to the reference dialogue history.
\textbf{(4) Model ranking score}: For each test question, the three memory modules—original memory bank, \ourmethod{}, and MemoRAG—are ranked based on response correctness. The models' scores are calculated as $s = 1/r, r\in\{1,2,3\}$, which indicates their ranking position.


\section{More Details About Reversible Transformers}
\label{appx:reversible_transformers}

In this section, we provide a more detailed explanation of reversible Transformers.

Reversible neural networks~\citep{dinh2014nice,dinh2022density} are constructed so that each layer's outputs suffice to exactly reconstruct its inputs. As shown in Figure~\ref{fig:reversible-nn}, a common paradigm is to split the input of layer $l$ into two groups, $x^1_l$ and $x^2_l$. Let the layer apply functions $\mathcal{F}_l$ and $\mathcal{G}_l$, producing outputs:
\begin{align}
    \mathbf{y}^1_{l}&=\mathbf{x}^1_{l}+\mathcal{F}_l\left(\mathbf{x}^2_{l}\right) \\
    \mathbf{y}^2_{l}&=\mathbf{x}^2_{l}+\mathcal{G}_l\left(\mathbf{y}^1_{l}\right)
\end{align}

Because $\mathbf{y}^1_{l}$ and $\mathbf{y}^2_{l}$ can be inverted as
\begin{align}
    \mathbf{x}^1_{l}&=\mathbf{y}^1_{l}-\mathcal{F}_l\left(\mathbf{x}^2_{l}\right) \\
    \mathbf{x}^2_{l}&=\mathbf{y}^2_{l}-\mathcal{G}_l\left(\mathbf{y}^1_{l}\right)
\end{align}

the forward transformation is bijective. Consequently, no intermediate activations beyond $\mathbf{y}^1_{l}$ and $\mathbf{y}^2_{l}$ need be stored in memory since intermediate states $\mathbf{x}^1_{l}, \mathbf{x}^2_{l}$ are fully recoverable in backward passes.

However, standard Transformers~\citep{vaswani2017attention} use residual connections and sub-layer stacks that do not conform to these precise invertibility requirements. To address this, \citet{liao2024make} propose making Transformer layers reversible with lightweight modifications. Each Transformer block is divided into two functional ``streams'': (1) the original sub-layer (attention or feed-forward), augmented with an adapter module~\citep{houlsby2019parameter}, and (2) a second stream that is an adapter-only module. The two streams act on two separate inputs $\mathbf{x}^1_{l}, \mathbf{x}^2_{l}$, arranged in a reversible pattern (analylogous to $\mathcal{F}_l$ and $\mathcal{G}_l$ above). One input passes through the original (frozen) Transformer sub-layer plus an adapter, while the other goes through a purely adapter-based function. The original Transformer parameters remain fixed, and only the adapters are trained. Because each stream can invert the other's output, the entire layer is fully reversible without re-training from scratch.

In our work, \ourmethod{} uses the reversible Transformer structure to simultaneously learn \emph{context compression} in the forward pass and \emph{context expansion} in the backward pass. By design, each layer allows exact reconstruction of its input. Thus, when we pass compressed representations ``backward'' through the network, we can recover the original text context. This bijective mechanism directly enforces consistency between memory retention (compression) and memory retrieval (expansion), enabling us to optimize both objectives together. Furthermore, exact invertibility minimizes activation storage and avoids the large overhead typically required to handle forward and backward passes in a standard Transformer.

\begin{figure*}[ht]
  \centering

  \begin{tcolorbox}[
      enhanced,
      segmentation hidden,
      colframe = gray!50!black,
      colback  = gray!10,
      boxrule  = 0.5pt,
      borderline={0.5pt}{0pt}{gray!50!black},
      sharp corners,
      fontupper=\ttfamily\small  
  ]
\textbf{Role:} You are an advanced language model specializing in hierarchical summarization.

\textbf{Task Overview}: Given a document, your goal is to decompose the content step by step into events, entities, sentences, and paragraphs. This process involves:

1. Identifying key events and the most relevant entities.

2. Gathering sentence-level contexts around those entities.

3. Constructing paragraph-level summaries from those sentence-level contexts.

4. Produce the final output as nested JSON that follows the structure.

\begin{enumerate}
  \item \textbf{Identify Events \& Select Entities}
  \begin{itemize}
    \item Scan the document to produce a set of query-worthy events.
    \item For each event, choose the key entities (people, places, organizations, and concepts) most relevant to that event.
  \end{itemize}

  \item \textbf{Gather Sentence-Level Context \& Form Sentence-Entity Pairs}
  \begin{itemize}
    \item For each entity, locate the sentences in which the entity appears or is crucially described.
    \item Create Sentence-Entity pairs: each pair references a sentence and the corresponding entity.
  \end{itemize}

  \item \textbf{Summarize into Paragraphs \& Link Document-to-Paragraph / Paragraph-to-Sentences}
  \begin{itemize}
    \item Group thematically related sentences together into paragraph-level summaries.
    \item Produce document-paragraph pairs and paragraph-sentence pairs.
  \end{itemize}

  \item \textbf{Final Output as Pure-String Hierarchical JSON}
  \begin{itemize}
    \item Maintain a strictly nested structure: 
    \item \texttt{Entities} must be a single string (e.g., ``EntityA,EntityB'').
    \item Ensure valid JSON syntax (quoted keys, values, and commas in the correct places).
  \end{itemize}
\end{enumerate}

\textbf{Example of Final JSON Structure (All Values as Strings):}

\begin{verbatim}
{
  "document_id": "d1",
  "paragraphs": [
    {
      "paragraph_id": "p1",
      "sentences": [
        {
          "sentence_id": "s1",
          "text": "Full text or condensed version of the first sentence.",
          "entities": "EntityA,EntityB"
        },
        {
          "sentence_id": "s2",
          "text": "Full text or condensed version of the second sentence.",
          "entities": "EntityA"
        }
      ]
    },
    {
      "paragraph_id": "p2",
      "sentences": [
        {
          "sentence_id": "s3",
          "text": "Full text or condensed version of the third sentence.",
          "entities": "EntityC,EntityD"
        }
      ]
    }
  ]
}
\end{verbatim}

  \end{tcolorbox}
  \caption{The prompt used to instruct GPT-4o to construct hierarchical context-query pairs in Section~\ref{sec:experiments}.}
  \label{fig:prompt}
\end{figure*}

\end{document}